# Semantically Guided Depth Upsampling

Nick Schneider*,1,3 Lukas Schneider*,1,2 Peter Pinggera[1]
Uwe Franke[1] Marc Pollefeys[2] Christoph Stiller[3]

* The first two authors contributed equally to this work
[1] Environment Perception, Daimler R&D, Sindelfingen, Germany
[2] ETH Zurich, Zurich, Switzerland
[3] Karlsruhe Institute of Technology, Karlsruhe, Germany
{nick,lukas}.schneider@daimler.com

**Abstract.** We present a novel method for accurate and efficient upsampling of sparse depth data, guided by high-resolution imagery. Our approach goes beyond the use of intensity cues only and additionally exploits object boundary cues through structured edge detection and semantic scene labeling for guidance. Both cues are combined within a geodesic distance measure that allows for boundary-preserving depth interpolation while utilizing local context. We model the observed scene structure by locally planar elements and formulate the upsampling task as a global energy minimization problem. Our method determines globally consistent solutions and preserves fine details and sharp depth boundaries. In our experiments on several public datasets at different levels of application, we demonstrate superior performance of our approach over the state-of-the-art, even for very sparse measurements.

## 1 Introduction

Many computer vision applications benefit from high resolution dense depth maps, e.g. image segmentation, scene flow computation, object detection and tracking, as well as 3D reconstruction and mapping. Traditionally, stereo vision has been the method of choice for obtaining per pixel depth estimates. However, accurate and dense state-of-the-art stereo algorithms[1] suffer from high computational complexity and distance-dependent measurement noise. In recent years, Light Detection and Ranging (LIDAR) sensors have become increasingly popular in the robotics domain due to their supreme range accuracy and constant improvements in size and cost. At the same time, compact and inexpensive Time of Flight (ToF) cameras have become key for indoor range sensing, providing per-pixel depth information at high frame rates. They however suffer from acquisition noise due to limited illumination energy and a comparatively low resolution compared to stereo camera systems. LIDAR sensors provide range measurements with very high accuracy in both indoor and outdoor scenarios, but only with a sparse and non-uniform sampling pattern.

[1] http://www.cvlibs.net/datasets/kitti/

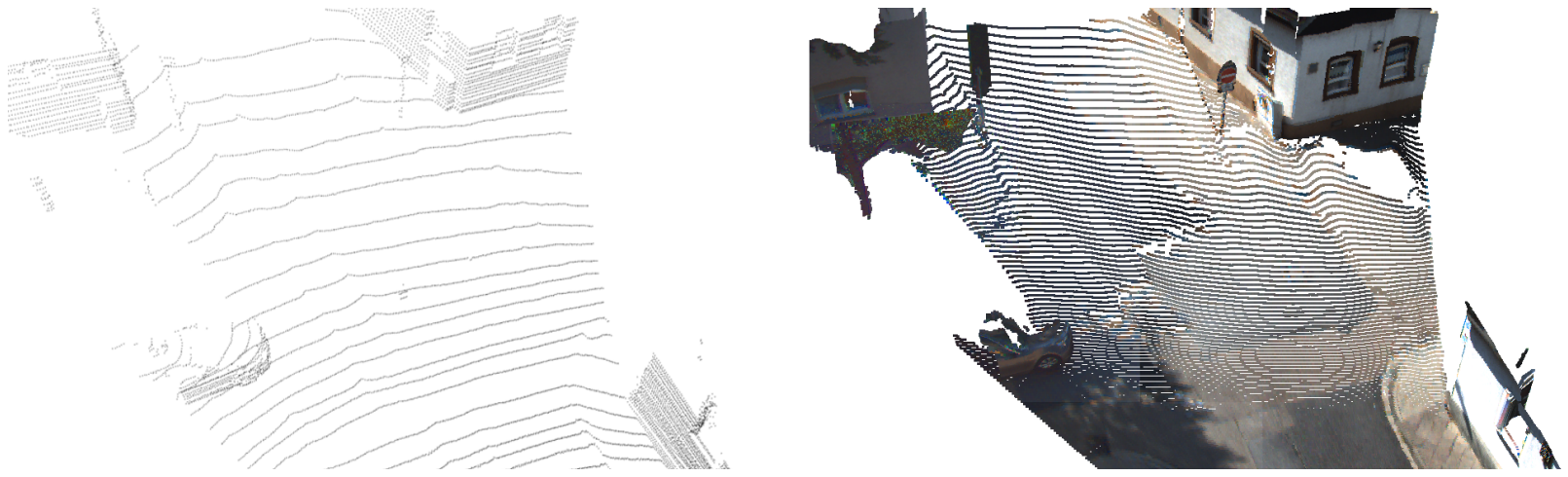

Fig. 1: Example output of the proposed method. The result is smooth and accurate, while fine structures, e.g. the pole and the sign, are preserved.

Sophisticated upsampling to a target resolution is a common method to address the sparseness of the measured depth data. Typically, a high resolution camera image is registered to the low resolution data and used to guide the upsampling process giving significant improvements, c.f. [10, 19, 25, 29]. The key assumption of such approaches is that depth boundaries coincide with intensity edges. We have however observed that actual object boundaries are more likely to induce depth discontinuities than plain intensity gradients. Hence, we aim to additionally exploit object boundary information as a guidance for upsampling.

In this paper, we present a novel image guided depth upsampling approach which utilizes both, object-based edge information as well as pixel-wise semantic class labels. For both cues, we leverage state-of-the-art approaches, i.e. structured edge detection [6] and pixel-wise semantic scene labeling [4]. The cues define a geodesic distance measure, allowing for boundary-aware distance computation between image pixels and sparse data samples. We model the observed scene structure as a set of locally planar elements and formulate the upsampling task as a global energy minimization problem. A thorough evaluation on several public datasets and different application levels demonstrates the benefits of our approach, particularly in terms of computing globally consistent solutions while preserving fine structures and sharp depth boundaries, c.f. Figure 1.

## 2 Related Work

Upsampling of sparse information through high resolution imagery is an active area in optical flow, stereo, and Time of Flight (ToF) research. All applications share two key assumptions: (1) the observed scenes consist of piecewise smooth surfaces, (2) depth and flow discontinuities tend to coincide with image edges. We identify four lines of related work:

First, many upsampling approaches apply an interpolation filter to the low resolution data. Since the filter is supposed to be guided by an associated high resolution image, the appropriate design of the filter kernel and its support is crucial. A popular approach involves bilateral filtering to combine sparse measurement data with ab high resolution guidance image [1, 7, 19, 36]. In [17], the guided filter is proposed as an efficient and improved alternative to bilateral upsampling. [26] exploit geodesic distances to define the filter support, achieving improvements at fine image details. Other methods rely on a previous segmentation of the image coupled with sophisticated smoothing operations [29, 30].

The second line of work formulates the upsampling task as a global optimization problem. [5] define a Markov Random Field (MRF), with the sparse depth measurements representing the data fidelity and the high resolution image serving as smoothness term. [32] incorporate adaptive multi-cue neighborhood weighting and nonlocal means regularization into the MRF to better preserve fine local details. [10] formulate a convex optimization problem with higher order regularization and an anisotropic diffusion tensor to handle fine structure. The higher order term enforces piecewise affine solutions and is modeled as a second order Total Generalized Variation (TGV) regularization. [9] compute geodesic neighborhoods and fit affine models to the low resolution data, with a subsequent greedy optimization scheme promoting global consistency. For optical flow upsampling [33] also use a geodesic distance representation to estimate the influence of sparse flow matches on an image pixel, followed by an energy minimization.

Third, a few approaches employ sparse signal representations for guided upsampling making use of the wavelet domain [16], learned dictionaries [23] or co-sparse analysis models [14, 18].

Finally, there are methods that estimate a dense depth representation leveraging pixel-level semantic cues as obtained by deep neural networks [2, 8, 24, 27, 28, 37]. Such methods either perform a joint inference on pixel-level using stereoscopic image data [3, 22], or operate on 3D reconstructed point clouds [11, 15, 20]. In this work, we also leverage semantic input through fully convolutional networks (FCNs) [28], but operate in the image domain and take sparse depth measurements as input. Our approach avoids common shortcomings such as over-smoothing and surface-flattening by combining a geodesic distance formulation with meaningful local geometric models and efficient global optimization.

Our main contributions are: (1) a novel energy formulation for depth upsampling that takes global image context into account, is robust to outliers and is optimized efficiently; (2) an extension of the well known geodesic distance to leverage probabilistic pixel-level semantic cues; (3) superior accuracy and computational requirements to state-of-the-art baselines.

## 3 Method

The goal of this work is to estimate a depth value for each pixel in a high resolution image, given sparse measurements. To describe an observed scene we infer planes, constrained to be consistent with their local context in the image. This formulation is driven by the observation that most object surfaces are intrinsically smooth and can thus be well approximated by local planes. We treat this task as a novel optimization problem of a global energy function, c.f. Section 3.1. In order to acquire meaningful context for each pixel, we leverage geodesic distances that respect image and semantic boundaries, c.f. Section 3.2. By approximating the geodesic distance, c.f. Section 3.3, we can drop most of the free parameters of the energy function and optimize it efficiently, c.f. Section 3.4.

### 3.1 Energy Formulation

Given a set $\mathcal{M}$ of pixels $i$ with sparse depth measurements $z_i$ and weights $w_{i,j}$ between $i$ and $j$, we formulate an energy function in order to estimate one plane $\theta_i \in \theta$ per pixel in the $u, v, \frac{1}{z}$ space as well as binary outlier indicators $\mathbf{o}$ for all measurements and co-planarity indicators $\mathbf{c}$ between all pairs of planes:

$$E(\theta, \mathbf{o}, \mathbf{c}, \mathbf{z}) = \sum_i E_{una}(\theta_i(i), o_i, z_i) + \sum_{i,j} w_{i,j} \cdot E_{pair}(\theta_i(i), \theta_j(i), c_{i,j}) + E_o(o_i, \theta, z_i, \mathbf{c}) \ . \tag{1}$$

If a measurement is marked as outlier, it is discarded in the unary that otherwise enforces consistency to its measurement:

$$E_{una}(\theta_i(i), o_i, z_i) = w_{una} \cdot \begin{cases} \left(\theta_i(i) - z_i^{-1}\right)^2 & \text{, if } i \in \mathcal{M} \wedge o_i = 0 \\ 0 & \text{, otherwise.} \end{cases} \tag{2}$$

The function $\theta_j(i)$ evaluates the inverse depth at pixel $i$ via $\theta_j \cdot [u_i, v_i, 1]^T$. In the pairwise term, the consistency of connected planes is reflected via

$$E_{pair}(\theta_i(i), \theta_j(i), c_{i,j}) = w_c \cdot \begin{cases} (\theta_i(i) - \theta_j(i))^2 & \text{, if } c_{i,j} = 1 \\ \lambda_c & \text{, otherwise,} \end{cases} \tag{3}$$

with a penalty $\lambda_c$ for non-consistent planes. Finally, the outlier formulation is kept as generic as possible by employing a probabilistic model taking the inferred planes into account:

$$E_o(o_i, \theta, z_i, \mathbf{c}) = \begin{cases} 0 & \text{, otherwise} \\ -log\left(p(o_i | \theta, z_i, \mathbf{c})\right) & \text{, if } i \in \mathcal{M}. \end{cases} \tag{4}$$

This potential is conveniently adaptable to the actual application demands and sensor properties. In this work, we opt for a simple outlier probability function that rates the measurement $z_i$ by judging its deviation from connected planes.

### 3.2 Geodesic distance

The proposed energy formulation depends on pairwise weights that control the influence of the pairwise energies. We define $w_{i,j} = -log\left(D_{i,j}\right)$ by means of a distance function $D_{i,j}$ between pixels in the image. With the intention of preserving depth discontinuities and tiny structures, we introduce an edge- and semantics-aware geodesic distance. It is defined as the costs along the minimal path $\pi$ from pixel $i$ to $j$. These costs are calculated as a weighted mean considering edge costs $s_I(\pi)$, semantic costs $s_S(\pi)$ as well as costs $s_l(|\pi|)$ which penalize the length of a considered path:

$$D_g(i, j) = \min_{\pi} \frac{w_I * s_I(\pi) + w_S * s_S(\pi) + s_l(|\pi|)}{w_I + w_S + w_D} \ . \tag{5}$$

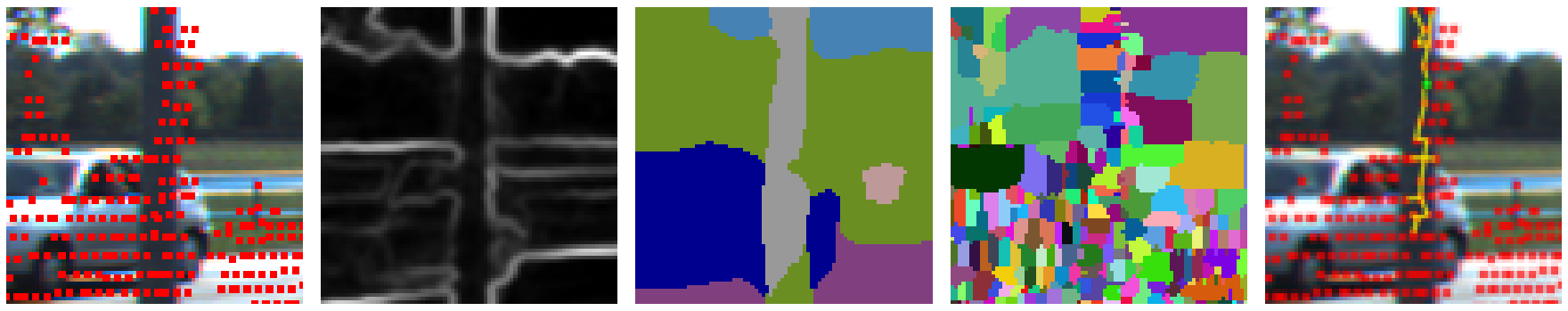

Fig. 2: The properties of the geodesic distance given the three inputs. The geodesic distance respects object boundaries as visible in the Voronoi and connected cells.

The edge term increases by traversing pixels with high edge scores $s_I(\pi) = \prod_{i \in \pi} s_I(i)$. Pixels along a path are more likely to be connected if they share the same semantic class. Therefore, we search for the label $l \in L$ which aggregates high scores $s_L(l, i)$ over all pixels $i$ along $\pi$. Since $s_L(l, i)$ increases with more certainty of label $l$, we define $s_S(\pi) = 1 - \underset{l}{\text{maximize}} \prod_{i \in \pi} s_L(l, i)$. Note that we expect normalized costs, thus $D_g(i, j) \in (0, 1] \Rightarrow w_{i,j} = -log(D_g(i, j)) \in \mathbb{R}^+$.

### 3.3 Approximation

We follow [33] and approximate the geodesic distance from Equation (5) with $\hat{D}_g(i, j)$. This allows us to efficiently compute the distances between all pixels in the image and simplifies the energy function. For each pixel $i$ we therefore firstly compute the nearest pixel $n_i \in \mathcal{M}$ with a valid measurement with respect to the original distance $D_g$ [33]. Using the measurement pixels as seeds, this step results in Voronoi cells $\mathcal{V}_i$, as shown in Figure 2. Now, we define $D_g^g(i, j) \approx D_g(i, j)$ by restricting the paths to contain the seeds in the traversed cells. This leads to $\hat{D}_g(i, j) = D_g(i, n_i) + D_g^g(n_i, n_j) + D_g(j, n_j)$. The distances to the closest pixel with valid measurement $D_g(i, n_i)$ are very small (assuming a reasonable number of measurements). Thus, we approximate them with a small constant $\epsilon$:

$$\hat{D}_g(i, j) \approx D_g^g(n_i, n_j) + \epsilon \; . \tag{6}$$

$D_g^g(n_i, n_j)$ can be computed efficiently using Dijkstra's algorithm, we refer to [33] for more details. The main advantage of the proposed approximation, however, lies in a significant reduction of the parameters necessary to optimize as we will show in the following. Let $i, j$ be pixels in the same cell $\mathcal{V}_k$. The approximated geodesic distance $\hat{D}_g(i, j) = \epsilon$ is very small leading to large pairwise weights $w_{i,j} = -log(\epsilon)$. The weights control the influence of the pairwise energy of Equation (1), thus forcing a solution of the energy function to satisfy $(\theta_i(i) \approx \theta_j(i))^2$. This leads to the observation that $\theta_i(i), \theta_j(i)$ have to intersect at $[u_i, v_i, \theta_i(i)]$. Since this observation holds for all pixel combinations in the cell, the plane parameters have to be equal: $\theta_i = \theta_j$ and we can replace all $\theta_i$ with $\theta_{n_i}$. In the following we denote the pixels with a valid measurement and their cell index as $n, m$. For pixels $i, j$ in two different cells with their corresponding measurement pixels $n, m$, the pairwise weights are constant and can be rewritten as $w_{n,m}$. We can thus reorder the summation and write

$$\sum_{i,j} w_{i,j} E_{pair}(\theta_i(i), \theta_j(i), c_{i,j}) = \sum_{n,m \in \mathcal{M}} w_{n,m} \sum_{p \in \mathcal{V}_n} E_{pair}(\theta_n(p), \theta_m(p), c_{n,m}) \quad (7)$$

Combining Equation (7) with the unary and outlier terms that evaluate to zero at pixels without valid measurements yields

$$E(\theta, \mathbf{o}, \mathbf{c}, \mathbf{z}) = \sum_{n \in \mathcal{M}} E_{una}(\theta_n(n), o_n, z_x) + \sum_{n,m \in \mathcal{M}} w_{n,m} \cdot \sum_{p \in \mathcal{V}_n} E_{pair}(\theta_n(p), \theta_m(p), c_{n,m}) + \sum_{n \in \mathcal{M}} E_o(o_n, \theta, z_n, \mathbf{c}) \ . \quad (8)$$

This new energy is effectively defined over cells instead of pixels reducing the number of free parameters in the optimization. However, given the reasonable approximation of the distance (and small $\epsilon$), both are essentially equivalent. In practice, we apply pairwise costs only to the $N$ nearest measurements with a maximal distance $\hat{D}_g < D_{max}$ by setting $w_{n,m} = 0$ otherwise. This results in a consistent set of cells allowing for robust depth optimization.

### 3.4 Optimization

Optimizing the energy function of Equation (8) is difficult, as it contains a large set of discrete and continuous variables. We adapt an iterative scheme from [35], repeatedly optimizing the coplanarity flags, outlier flags and plane parameters one after another. The energy is convex in $\theta$ and can efficiently be solved in closed form. It is carried out in parallel for all planes, assuming the rest of the planes and parameters as given, using polynomial coefficients to evaluate the unary and pairwise potentials in constant time. Although there is no guarantee for the energy to decrease in each iteration, in our experiments the measured depth error decreases consistently with an increasing number of iterations, c.f. Section 4. In such an iterative optimization scheme, good results are often extremely sensitive to the initialization. Throughout our experiments, however, we opt for a simple initialization by $o_i = 0, c_{i,j} = 1$ and $\theta_i = [0, 0, z_i^{-1}]$. With available label input, we greedily decide if a cell is dominated by ground or object pixels independently. In the former cases, we initialize the planes with upright normals instead.

## 4 Evaluation

We evaluate our approach on three different application levels. First, we use the sampled groundtruth of the Middlebury 2014 stereo dataset [34]. Second, we evaluate our algorithm on two real-world datasets: the indoor dataset introduced by Ferstl [10] and the KITTI dataset ([13], [12] and [31]). We opt for a generic and efficient edge detector [6] that uses trained structured forests to predict edge scores for each pixel.

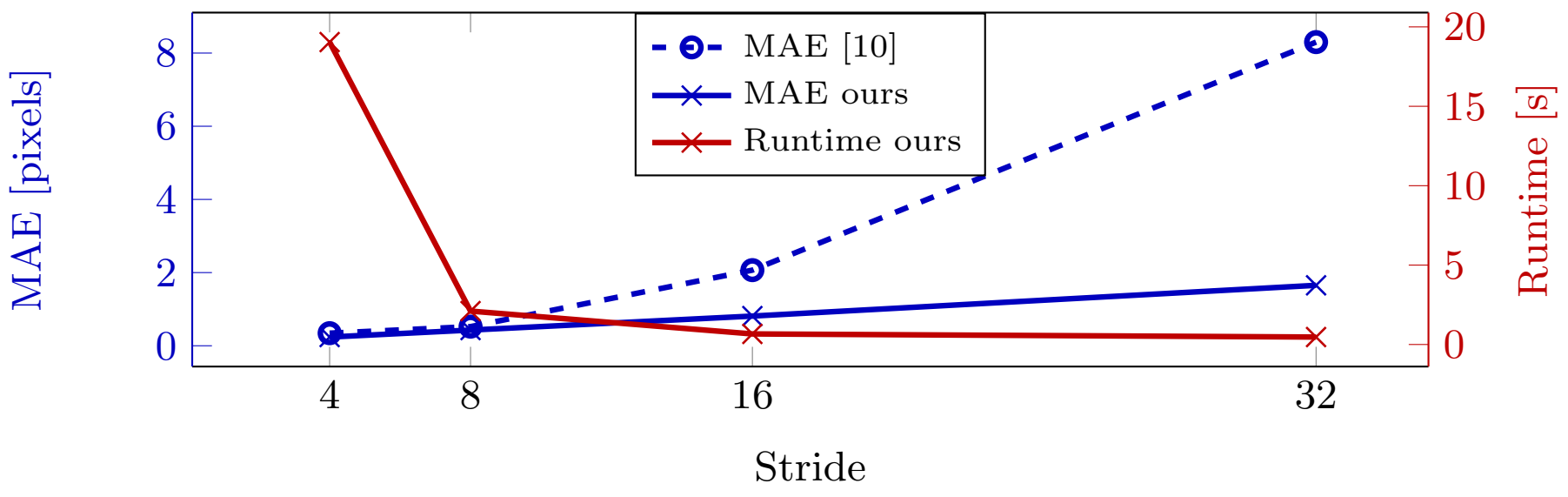

Fig. 3: Comparison of runtime and MAE (in pixel) for different strides on the Middlebury 2014 Benchmark [34].

### 4.1 Middlebury Benchmark Evaluation

To test the performance of our algorithm on almost noise-free input data, we make use of the recent Middlebury 2014 Stereo Benchmark [34]. The dataset contains high resolution images and corresponding groundtruth generated by structured lighting. Previous depth super resolution methods [10, 32] used Bicubic or Bilinear downsampling to simulate the input. As this distorts the input data especially at depth edges and blurs tiny structures, we simply sample data at equidistant points. We compare to a state state-of-the-art method of [10] and three additional upsampling approaches. The parameters of our and Ferstl's [10] algorithm were empirically determined for each upsampling factor. Quantitative results in terms of the Mean Absolute Error (MAE) are shown in Table 1. All baselines are significantly outperformed on all three images and all strides, particularly for highly sparse inputs. The runtime of our algorithm, visualized in Figure 3, strongly depends on the number of input depth points. While [10] report moderate runtimes of their algorithm (318.2 ms for upsampling 0.8 megapixel images and 1900 ms for 1.4 megapixel images), the speed depends on the resolution of the image instead. Thus, our algorithm is best suited for real-world depth sensors, which so far have a very limited resolution. Figure 3 visualizes the performance and runtime of our method in comparison to the number of input measurements.

Table 1: MAE for Middlebury 2014 dataset for four chosen images.

| | Backpack | | | | Sword1 | | | | Umbrella | | | |
|---|---|---|---|---|---|---|---|---|---|---|---|---|
| | 4x | 8x | 16x | 32x | 4x | 8x | 16x | 32x | 4x | 8x | 16x | 32x |
| Nearest | 2.65 | 3.95 | 5.24 | 6.74 | 1.72 | 2.94 | 4.65 | 7.16 | 0.77 | 1.26 | 2.18 | 4.06 |
| Bicubic | 2.22 | 3.44 | 4.76 | <u>6.07</u> | 1.44 | 2.56 | 4.23 | <u>6.65</u> | 0.67 | 1.10 | 1.90 | 3.64 |
| Bilinear | 1.95 | 3.21 | 4.56 | 6.20 | 1.28 | 2.40 | 4.12 | 6.75 | 0.56 | 0.94 | 1.69 | <u>3.37</u> |
| Ferstl | <u>0.50</u> | <u>0.79</u> | <u>3.37</u> | 11.54 | <u>0.49</u> | <u>0.75</u> | <u>2.61</u> | 9.43 | <u>0.15</u> | <u>0.22</u> | <u>1.37</u> | 9.10 |
| OURS | **0.15** | **0.27** | **0.47** | **0.90** | **0.34** | **0.61** | **1.07** | **2.22** | **0.09** | **0.16** | **0.27** | **0.53** |

Table 2: Results on the Ferstl Dataset [10] by means of MAE error in depth [mm].

| | Books | Shark | Devil |
|---|---|---|---|
| Kopf [19] | 16.03 | 18.79 | 27.57 |
| He [17] | 15.74 | 18.21 | 27.04 |
| Ferstl [10] | 12.36 | **15.29** | 14.68 |
| OURS | **12.12** | 15.46 | **14.03** |

### 4.2 Results for ToF data upsampling

The Middlebury dataset is popular for the evaluation of depth upsampling methods but does not reflect real acquisition setups. Therefore, we further carry out an exhaustive evaluation on two real-world datasets. The first is provided by [10] and consists of three different scenes captured by a 120×160 pixel wide ToF camera and a 810×610 pixel wide greyscale CMOS camera. Groundtruth for these scenes was generated using a structured light scanner with a high-speed projector as well as two 2048×2048 pixel wide high-speed intensity cameras (with a depth uncertainty of 1.2 mm). We compare our results to three upsampling algorithms: joint bilateral upsampling [19], guided image filtering [17] and anisotropic total generalized variation [10]. A quantitative comparison of the results is shown in Table 2, qualitative results in Figure 4. We demonstrate that our method produces smooth surfaces while preserving sharp depth edges and thin elements. The runtime for a single frame is 107 ms on the given input data and was measured as an average over 1000 runs (Ferstl [10] reports an average of 318.2 ms).

### 4.3 Results for LIDAR upsampling

The KITTI Vision Benchmark Suite [13] was created to fill the gap of demanding benchmarks for visual systems in the context of autonomous driving. The dataset provides depth data from a Velodyne HDL-64E LIDAR as well as RGB images with a resolution of 1392×512 pixels which were recorded from a moving platform. With the help of a highly accurate inertial measurement unit (IMU) LIDAR depth data is accumulated while driving and serves as groundtruth for stereo vision methods. In [31], a novel dataset was presented in which also frames with moving objects are considered. The dynamic objects are first removed and then re-inserted by fitting CAD models to the point cloud. Furthermore, occlusions due to sensor displacement are manually removed, resulting in a clean and dense ground truth for depth evaluation. We use the raw LIDAR measurements as input to our algorithm. It is apparent from Figure 6 that the groundtruth differs extremely from raw input in terms of both, outlier frequency and density. We traced the raw LIDAR data associated to the training images of the 2015 Stereo Benchmark and found 82 corresponding frames from the raw dataset, which we will make publicly available to allow for future evaluation on this data. The raw LIDAR points projected into the image $(u, v, d)$ contain many errors due to the large displacement between the LIDAR and the camera especially at top borders of objects. Instead of heuristically removing measurements, we aim for an outlier rejection in the proposed optimization, c.f. Section 3. We tackle this challenge by leveraging pixel-level semantic information in addition to the sparse depth measurements. Therefore, we use a FCN [28] trained on the Cityscapes Dataset [4] and fine-tune it on a disjunct part of the KITTI images annotated with pixel-wise labels [21]. It is trained with a label set consisting of 11 labels including road, sidewalk, car, pole and building. We report the MAE over all pixels having a groundtruth disparity. In order to evaluate the performance of the proposed method and the baseline with different input densities, we down-sample the vertical and horizontal resolution of the LIDAR output (Figure 5). Our method significantly outperforms all baselines, while the performance gap grows for sparser input data. In particular the semantic cues play a greater role with less measurements. For the baseline [10], we optimized the parameters to the best of our knowledge using all 82 frames of the evaluation.

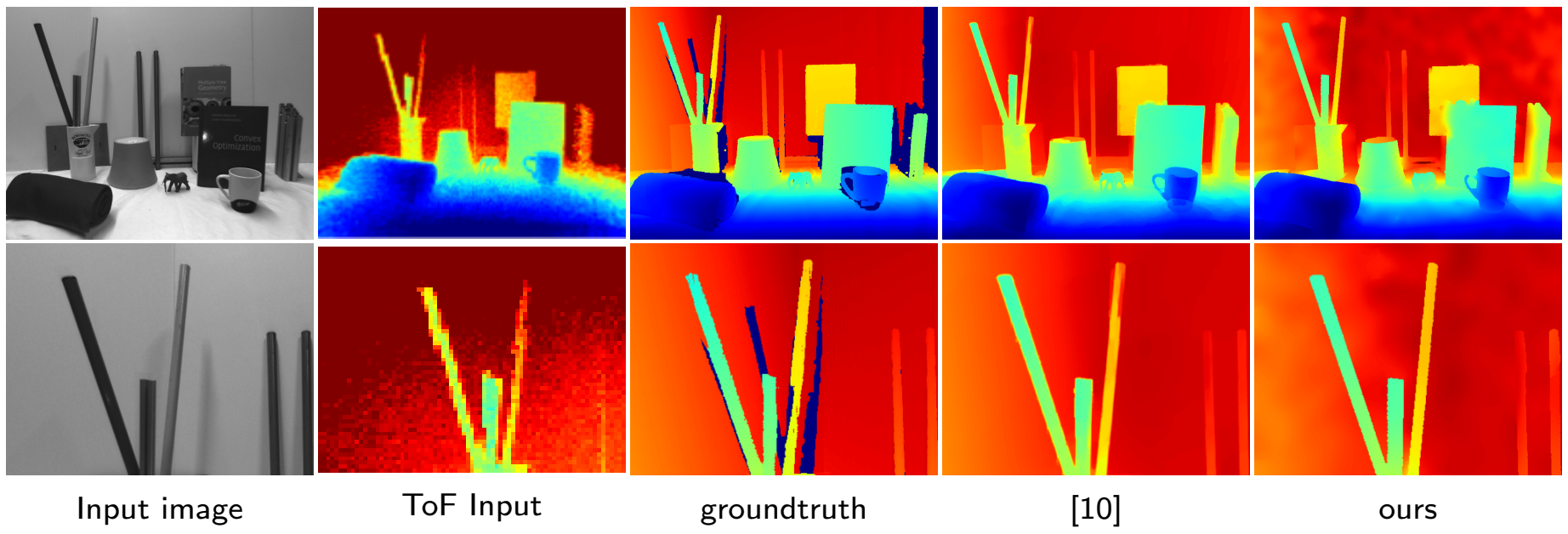

Fig. 4: Result for upsampling the Time of Flight data on the ‘books’ scene of the Ferstl dataset [10]. In the second row a zoomed-in view is shown for better visual comparison. Although input data is noisy, our algorithm preserves sharp edges.

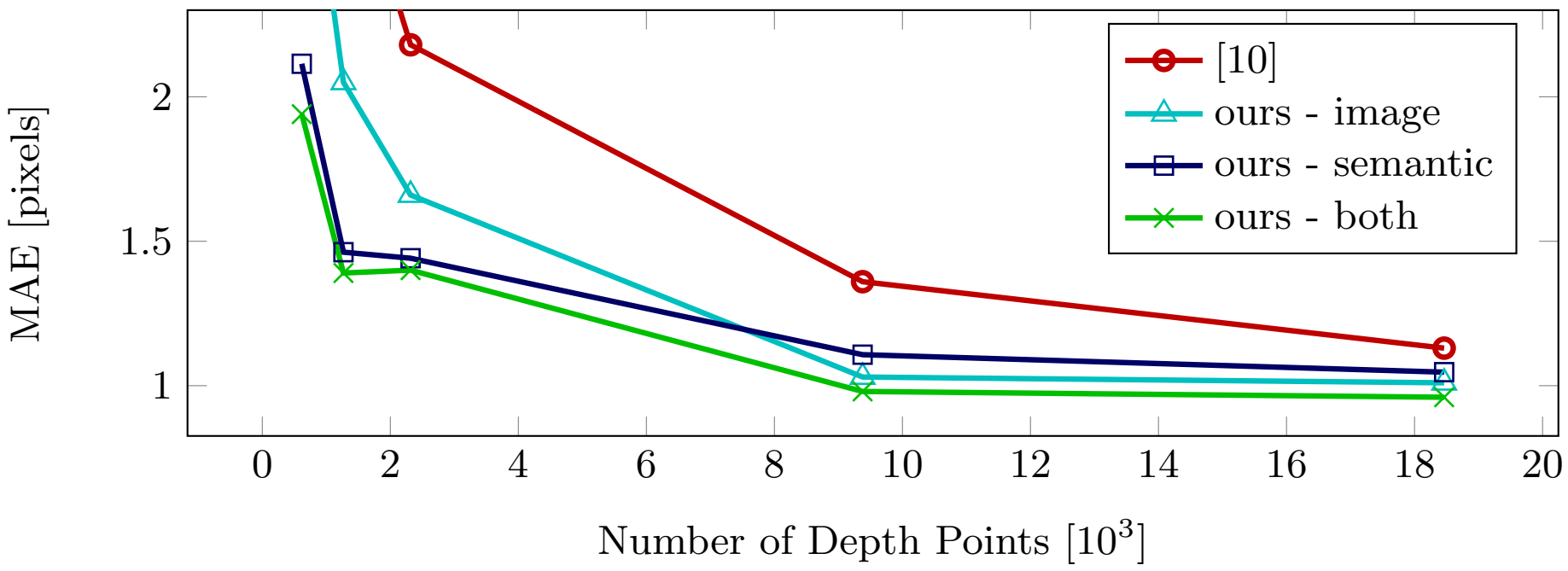

Fig. 5: Quantitative results on the Kitti dataset for varying number of input measurements. Results in terms of disparity MAE depending on the number of input points are reported for a baseline [10] and three input configurations (image only, semantic only, both inputs) of the proposed method.

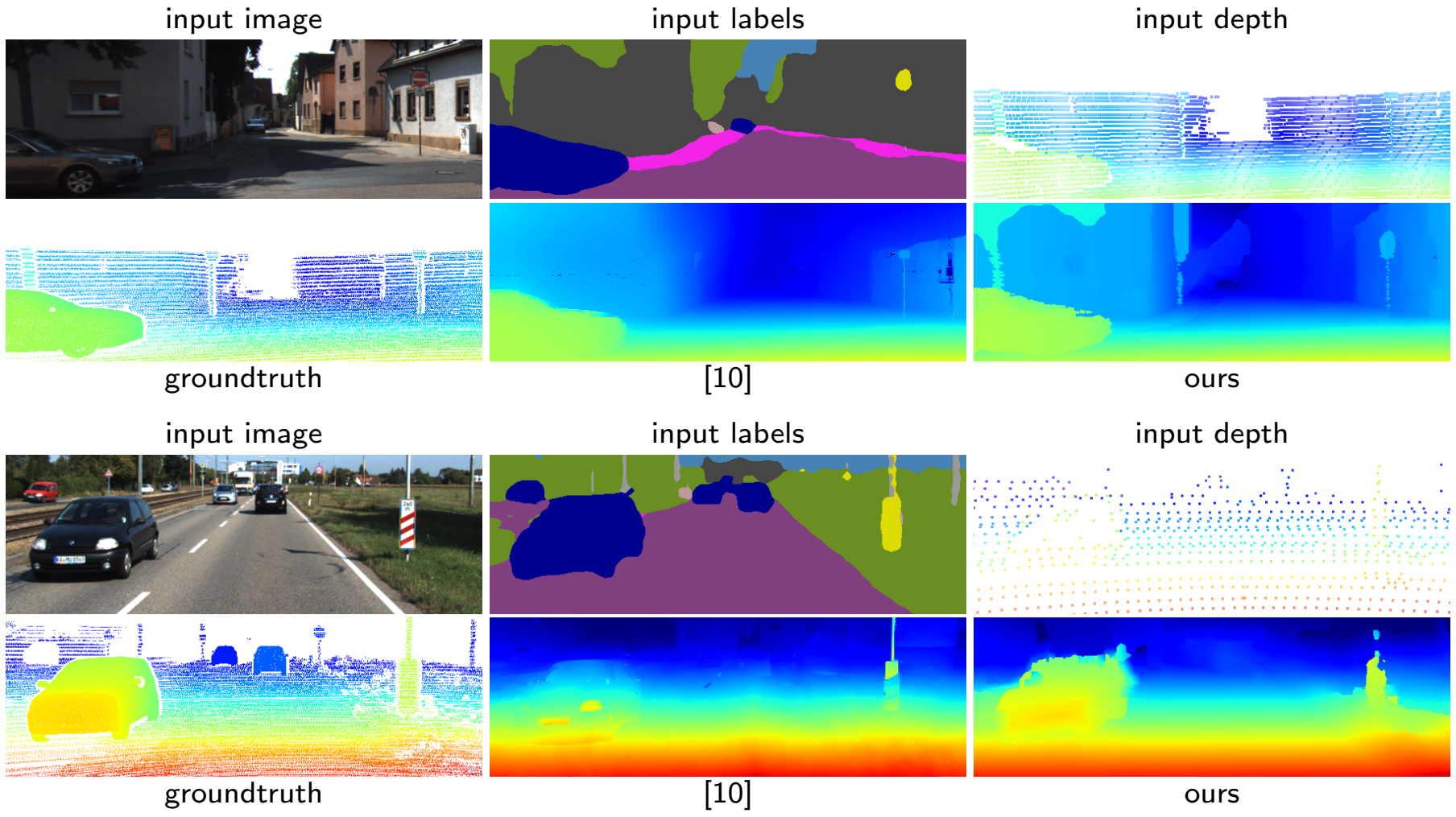

Fig. 6: Example outputs of the proposed algorithm compared to our baseline, groundtruth and the inputs for two cases with all (top) respectively $\frac{1}{12}$ (bottom) measurements. The semantic input guides our upsampling to respect object boundaries as can be seen at the objects, e.g. car, signed and poles, in the lower example. The 3D reconstruction using the inferred depth is presented in Figure 1. Note that the sign on the right (top) is correctly represented without available depth measurements, by propagating the depth along the sign and pole.

## 5 Conclusion

In this paper, we presented a novel approach for the challenging problem of accurate depth upsampling. Our proposed method exploits boundary cues and pixel-wise semantic class labels obtained via a high resolution guidance image and fully convolutional networks. In order to preserve depth boundaries and fine structures, we combine those cues in a geodesic distance measure. We formulate the upsampling task as a pixel-wise global energy minimization problem and apply a suitable approximation which allows to reduce the number of parameters for a real-time optimization. A thorough evaluation on three different public datasets is carried out at different application levels. Compared to a state-of-the-art method, we achieve significantly better results. The proposed method can particularly exploit its strength at very sparse depth measurements or a high target resolution. Here, the performance gap to the baselines increases in terms of both, computational demands and depth accuracy. Furthermore, the experiments demonstrated robustness of our approach to noise and false depth measurements.